\documentclass[letterpaper]{article}
\usepackage{aaai20}
\usepackage{times}
\usepackage{helvet}
\usepackage{courier}
\usepackage[hyphens]{url}
\usepackage{graphicx}
\usepackage{todonotes}
\usepackage{esvect}
\usepackage{latexsym}
\usepackage{mathtools}
\usepackage{tcolorbox}
\usepackage{breqn}
\usepackage{graphicx}
\usepackage{multirow}
\usepackage{amsmath}
\usepackage{amssymb}
\usepackage{todonotes}
\usepackage{subcaption}
\usepackage{url}
\usepackage{booktabs}
\usepackage{array}
\usepackage{threeparttable}

\DeclareMathOperator{\E}{\mathbb{E}}
\newcommand{\tabhead}[1]{\textbf{#1}}

\usepackage{array}
\usepackage{placeins}
\newcolumntype{?}{!{\vrule width 1pt}}
\title{A Comparison of Architectures and Pretraining Methods for Contextualized Multilingual Word Embeddings}
\author{Niels van der Heijden\\
ILLC, University of Amsterdam\\{ n.vanderheijden@student.uva.nl}
\And Samira Abnar\\
ILLC, University of Amsterdam\\{ s.abnar@uva.nl}
\And Ekaterina Shutova\\
ILLC, University of Amsterdam\\{ e.shutova@uva.nl}}

\begin{document}
\maketitle
\begin{abstract}
    The lack of annotated data in many languages is a well-known challenge within the field of multilingual natural language processing (NLP). Therefore, many recent studies focus on zero-shot transfer learning and joint training across languages to overcome data scarcity for low-resource languages. In this work we (i) perform a comprehensive comparison of state-of-the-art multilingual word and sentence encoders on the tasks of named entity recognition (NER) and part of speech (POS) tagging; and (ii) propose a new method for creating multilingual contextualized word embeddings, compare it to multiple baselines and show that it performs at or above state-of-the-art level in zero-shot transfer settings. 
    Finally, we show that our method allows for better knowledge sharing across languages in a joint training setting.
\end{abstract}

\section{Introduction}
With the aim of extending the global reach of NLP technology, much recent research has focused on the development of multilingual models. Due to the lack of annotated data and resources in many languages, these models typically rely on multilingual joint learning or zero-shot transfer learning across languages. Much of this research has utilized multilingual word embeddings \cite{muse,vecmap}, that project words in multiple languages into a shared multilingual semantic space, such that translations of words across these languages appear close in the new space. Such general-purpose multilingual word embeddings then serve as a basis for joint or transfer learning focused on a particular task.

On the other hand, current best performing monolingual approaches have moved away from static to contextualized word embeddings. Such contextualisation allows the models to address the long-standing problem of polysemy and dynamically model contextual meaning variation. The first such model, CoVe \cite{cove}, used a deep LSTM \cite{lstm} encoder pretrained in a machine translation task to contextualize word vectors. This work paved the way for contextualized word embeddings by showing their superiority over the static ones on downstream tasks such as named entity recognition (NER) and question answering. Shortly after, ELMo \cite{elmo} improved upon CoVe by making the contextualized embeddings \textit{deep} by combining information from multiple layers of their LSTM encoder trained with language modelling objective.
Finally, the Transformer-based \cite{vaswani2017attention} BERT model \cite{bert} broke the performance records across many downstream NLP tasks, achieving up to 7.6\% absolute improvement over the previous state of the art. 

However, in the multilingual setting, the effects of contextualization are still relatively unexplored. One exception includes the work of Schuster et al, \shortcite{schuster2019cross} that generalizes the work of Lample et al, \shortcite{muse} to the ELMo model by viewing a contextual word embedding as a context-dependent shift from the (static) mean embedding of a word. The mean embeddings are aligned across languages using adversarial training. This technique, however, does not outperform the non-contextualized embedding baseline in half of the performed experiments when no supervised anchored alignment is given and does not allow for large-scale joint training across languages.

Meanwhile, recent multilingual NLP research has moved away from solely word-level representations to training ``massively'' multilingual sentence encoders. Artetxe and Schwenk, \shortcite{laser} proposed a method for learning language-agnostic sentence embeddings (LASER) for 93 languages with a single shared encoder and limited aligned data. More specifically, they trained a deep LSTM-encoder to embed sentences in all (93) languages into a shared space such that semantically similar sentences in different languages appear close to each other in this space. Using this model the researchers advanced the state of the art on zero-shot cross-lingual natural language inference for 13 out of 14 languages in the XNLI \cite{xnli} dataset. Following these advances, Lample and Conneau, \shortcite{xlmberts} incorporated multilingual language pretraining into the BERT model \cite{bert}, coining their model XLM (cross-lingual language model), and again further improved the performance on all XNLI languages. Such multilingual sentence encoders have so far been applied and evaluated in sentence-level tasks. Yet, they provide a promising starting point for learning multilingual contextualized word representations needed for word-level classification tasks 
and investigating the effects of contextualisation in multilingual word-representation models.

The contributions of this paper are as follows: 
\begin{itemize}
     \item We conduct a comprehensive comparison of state-of-the-art multilingual word and sentence encoding models and pretraining methods on the tasks of NER and part of speech (POS) tagging, experimenting in both zero-shot transfer learning and joint training setting.
    \item We introduce a new method for learning contextualized multilingual word embeddings based on the LASER encoder and perform an in-depth analysis on its performance against multiple benchmarks in zero-shot transfer and joint training settings. We improve the previous state-of-the-art for English to German NER with 2.8 F1-points and perform at state-of-the-art level for other languages.
    \item We empirically show and analyze the benefit of contextual word embeddings versus static word embeddings for zero-shot transfer learning.
\end{itemize}

\section{Related work} \label{sec:rel work}

\subsection{Monolingual word representations}
After the success of static word embeddings such as  word2vec \cite{word2vec} which produced general-purpose semantic encodings of words, a lot of effort has been put into contextualizing these embeddings. While static embeddings at the time improved results when applied to various NLP tasks, polysemy has remained a challenge for these models, as they represent all meanings of a word within one vector. 
All occurrences of a word are thus treated the same and the resulting vector is a combination of the semantics of each possible meaning of the word.

In order to incorporate the context into embeddings, Peters et al. \shortcite{bilstmLM} introduced \textit{Language Model (LM) embeddings} and showed that they improved sequence tagging performance, specifically for NER. Subsequently, Peters et al. \shortcite{elmo} took the LM embeddings a step further by making them deep, which resulted in performance improvements in several downstream benchmarks. Their model, ELMo, incorporates information from all layers of the network by taking a layer-wise weighted average of the embeddings. Interestingly, the authors showed that each layer of their LSTM-encoder encoded different properties of the word. The first layer captures more syntactic aspects of a word whereas the second layer captures more high-level semantic information. 

\subsection{Multilingual word representations}
Much previous research has focused on aligning word embeddings from different languages into a language independent space  \cite{ap2014autoencoder}, either bilingual or multilingual. These methods either jointly train word embeddings on aligned corpora or align monolingual ones by means of post-processing. An obvious drawback of these methods is their need for aligned corpora. Hence in more recent work the focus shifted towards unsupervised alignment of word embeddings. 
Works such as 
Multilingual Unsupervised or Supervised word Embeddings (MUSE) \cite{lample2017unsupervised} is an example of unsupervised methods capable of aligning embeddings into a shared space, enabling easier knowledge transfer across languages without the need for additional resources. By aligning the embedding spaces of more than 30 languages, it generates high-quality embeddings for use in multilingual semantics tasks. Because of its proven performance we use these embeddings as a baseline to compare our models to. 

To the best of our knowledge, the work of Schuster et al. \shortcite{schuster2019cross} comes closest to ours. They present a method to align monolingual ELMo embeddings across languages by modelling such an embedding as a context-dependent shift from its mean (see equation \ref{eq:mean_embed}, where $e_i$ is the embedding for the work $i$ and $c$ is the context) and applying the linear alignment technique proposed by Mikolov, Le and Sutskever \shortcite{mikolov2013exploiting} to the mean embedding of each word.
\begin{align}
    e_{i,c} = \overline{e}_i+\hat{e}_{i,c}, \qquad \overline{e}_i = \E_c[e_{i,c}] \label{eq:mean_embed}
\end{align}
Whereas the authors show this approach to be a simple yet effective one, it does not allow for joint training across languages or handling code-switching. Our proposed method tries to overcome these deficiencies by sharing one encoder for all languages instead of transferring knowledge from one to another language by means of post-processing.

\subsection{Zero-shot cross-lingual transfer learning}
Early work on cross-lingual transfer learning used parallel corpora to create cross-lingual word clusters or exploited external knowledge bases as means of feature engineering \cite{tackstrom2012cross}. More recent approaches either exploit bilingual word embeddings to translate a dataset \cite{xie2018neural} or attempt to learn language-invariant features \cite{chen2019multi}.
\subsubsection{LASER}
A related promising development is that of Language-Agnostic SEntence Representations (LASER) \cite{laser}, where one of the main contributions is an encoder capable of embedding sentences in 93 languages in a shared space such that semantically related sentences are close in this space, regardless of their respective language, language family and script. At the time of release, LASER has set a new state of the art on multiple zero-shot transfer learning tasks such as XNLI \cite{xnli}, indicating its success in creating language agnostic embeddings. 

The LASER encoder consists of a byte-pair encoded vocabulary (BPE) \cite{bpe} followed by a 5-layer biLSTM with 512-dimensional hidden states. The final sentence embedding is obtained by applying max pooling over the hidden states of the final layer. BPE is a form of learning Subword Units (SUs) by encoding frequently occurring character n-grams as a symbol. In the case of LASER 50k of those symbols were learned together with their respective embedding.

\begin{figure}
    \centering
  \includegraphics[width=\linewidth]{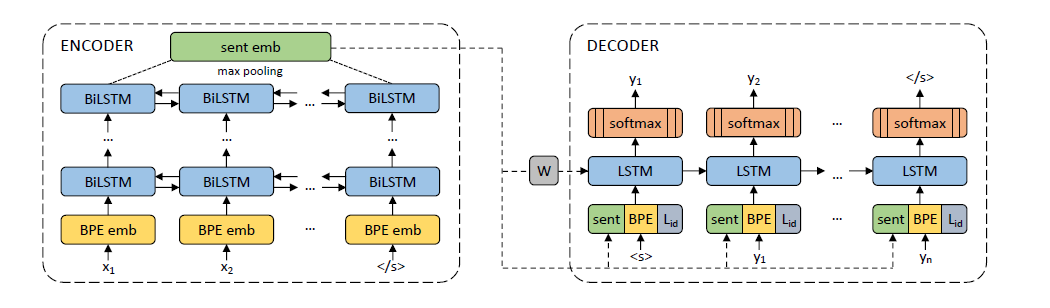}
  \caption{Encoder-decoder setup for training LASER}
  \label{fig:laser}
\end{figure}

The encoder is trained in an encoder-decoder setup in the task of machine translation, as shown in Figure \ref{fig:laser}. More specifically, a dataset was gathered by combining the Europarl, United Nations, Opensubtitles2018, Global Voices, Tanzil and Tatoeba corpora \cite{tiedemann2012parallel} comprising sentences in 93 languages translated into English and/or Spanish. The task of the encoder is to encode a sentence in a 1024-dimensional vector such that the decoder can generate the translation of the original sentence in a chosen target language. The decoder receives no information about which language is encoded by the encoder and hence it cannot distinguish between languages. This forces the encoder to create language-agnostic sentence embeddings. 

The main contribution of LASER is that the the authors show that it is possible to encode numerous languages into one encoder when a shared vocabulary is learned and the training data is aligned with just two target languages. 
Since the LASER sentence encoder achieves very promising results on zero-shot transfer learning for sentence-level NLP tasks, we will use this model as a basis for ours. 
Our goal is to investigate the possibility of extracting contextualized word embeddings from an encoder trained at the sentence level. We evaluate two versions of our model and compare them to multiple baselines.

\subsection{Multilingual joint learning}
Multilingual joint learning has shown to be beneficial when either the target or all languages are resource-lean \cite{khapra2011together}, when code-switching is present \cite{adel2013combination} or even in high-resource scenarios \cite{mulcaire2018polyglot}. Often, multilingual joint training is approached by some form of parameter sharing \cite{johnson2017google}.

\section{Methods} 
\begin{figure}
    \centering
  \includegraphics[width=\linewidth]{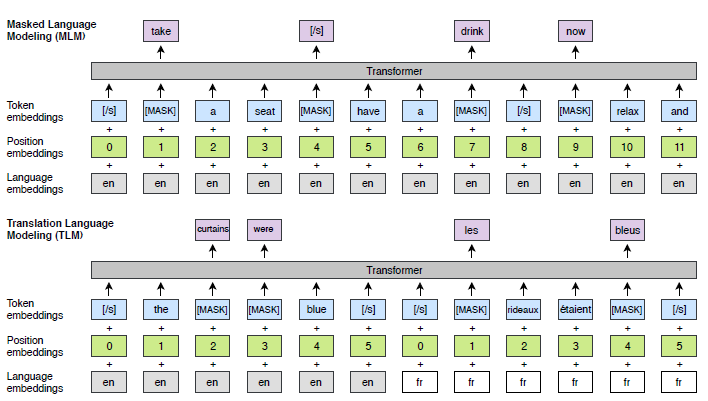}
  \caption{Masked Language Model and Translation Language Model}
  \label{fig:mlm_tlm}
\end{figure}
\subsection{LASER-based contextualized embeddings} \label{sec:prop_method}

We use a pretrained LASER model to obtain the contextualized word embeddings. For this, we compare different methods, which we explain below.

\noindent \textbf{BPE BOW} \qquad As the first baseline, we simply create word embeddings by averageing the BPE embeddings per word. This approach can be compared to a continuous Bag-Of-Words (BOW) approach with the BPE embeddings serving as words.
 
\noindent \textbf{BPE GRU} \qquad As the second baseline, we introduce a GRU \cite{gru} encoder followed by max-pooling over time to encode the BPE embeddings into a word embedding. First, each BPE symbol is embedded by the pretrained embeddings from the LASER encoder. Then, the data is split into a [BPE\_pad, N, Emb\_dim] tensor with BPE\_pad the length of the longest word in the batch expressed in number of Subword Units. N is the number of words in the batch and Emb\_dim is the embedding dimension of the pretrained BPE symbols which equals to 320. This tensor is then fed into the GRU encoder to compute the semantics of the SUs together. The final embedding is created by applying max-pooling over time on the output of the GRU.

\noindent \textbf{MUSE} \qquad As the third baseline, we consider static crosslingual word embeddings from the MUSE model \cite{lample2017unsupervised} as embeddings for our sequence tagger. 

\noindent \textbf{LASER-top} \qquad Our first proposed method incorporating the LASER LSTM encoder, which we call \textit{LASER-top}, uses the hidden state of the final layer as a base representation of a BPE symbol. First, we apply max-pooling over the forward and the backward hidden states, $\overrightarrow{ \mathbf{h}}^{(t)}_L$ and $\overleftarrow{ \mathbf{h}}^{(t)}_L$, to obtain a 512-dimensional vector per BPE symbol\footnote{Preliminary experiments showed that max-pooling both reduced overfitting and improved computational efficiency.}. Inspired by ELMo, we then rescale this output with a learnable scale parameter $\gamma$.
\begin{align}
    \mathbf{m}^{(t)}_L = \gamma \cdot maxpool(\overleftarrow{ \mathbf{h}}^{(t)}_L, \overrightarrow{ \mathbf{h}}^{(t)}_L),  L=5 \label{eq:scaled_mp}
\end{align}
As the LASER encoder is fed a sentence split into SUs its output can now be seen as a contextualized representation of the original embeddings, which is the reason why we expect this method to improve over the baselines.
Then, similar to the original approach \cite{laser}, the final word embedding is created by applying max-pooling over time to all $\mathbf{m}^{(t)}_L$s belonging to a word, for each word separately. 

\noindent \textbf{LASER-elmo} \qquad Inspired by ELMo and hence called \textit{LASER-elmo}, we make our multilingual contextualized embeddings \textit{deep} by incorporating multiple layers of the LSTM encoder. In order to do this, a weighted average of the hidden states of all layers is computed by softmax-normalizing task-specific layer weights $\mathbf{s_l}$ for layer l, which are learned during training.
\begin{align}
    \mathbf{e}^{(t)} = \Sigma_{l=1}^5 \mathbf{s_l} \cdot \mathbf{m}^{(t)}_l
\end{align}
Where $\mathbf{e}^{(t)}$ is the deep contextualized embedding for the SU at index $t$ in the sequence, $\mathbf{m}^{(t)}_l$ is computed as in Equation \ref{eq:scaled_mp} and $s_l$ is the softmax-normalized layer weight. These embeddings are then used as in LASER-top to create word embeddings. \\
\textbf{LSTM No Pretraining} \qquad In order to verify what part of the performance of LASER-top and LASER-elmo can be attributed to the fact that the 5-layer LSTM encoder allows for modelling more complex dependencies, we replace the pretrained encoder by a randomly initialized one and retrain. As overfitting plays a major role in training our models and Peters et al, \shortcite{elmo} have shown a two-layer LSTM to be sufficiently powerful for sequence tagging tasks, we pick a two-layer LSTM to replace the LASER encoder. Otherwise, this baseline functions in the same way as LASER-elmo.

\subsection{Transformer-based contextualized embeddings}
\textbf{BERT} \qquad Devlin et al. \shortcite{bert} propose a novel language representation model which uses Transformers to create deep contextual representations for words. These representations are obtained by training the model on unlabeled text to predict the words at randomly chosen masked positions conditioned on both the left and the right context - the authors call this technique the Masked Language Model (MLM), see Figure \ref{fig:mlm_tlm}. BERT uses WordPiece embeddings \cite{wu2016google} with a vocabulary of 30k tokens and is trained on the BookCorpus \cite{zhu2015aligning} and a Wikipedia dump together: a combined corpus of approximately 3300M words. In addition to the MLM, BERT is also trained on a next sentence prediction task in order to capture relationships between sentences. This task is phrased as a binary classification task where either two consecutive or two random sentences are sampled from the corpus. The complete pretraining procedure combines these two tasks by sampling sentences as described for the next sentence prediction task and applying both this task and the MLM.

Although BERT is not explicitly pretrained to align semantics across languages, its multilingual version\footnote{The multilingual version of BERT is not described in the original paper by \cite{bert}. Instead, it is described on the official GitHub page of the authors: \url{https://github.com/google-research/bert/blob/master/multilingual.md}.}, from which we use the cased base version in our experiments, is trained on 100+ languages and its monolingual capabilities are (near) state-of-the-art for many NLP tasks without heavily-engineered task-specific architectures. 

\noindent \textbf{XLM} \qquad After the success of LASER in zero-shot multilingual transfer learning Lample and Conneau \shortcite{xlmberts} proposed a similar method based on the architecture of BERT. Their contribution lies in the introduction of several new unsupervised and supervised methods for cross-lingual language model pretraining. In this work we will focus on their supervised method as it is most closely related to LASER and outperforms the former on the XNLI benchmark. This method is a multi-task setup of a slightly adjusted MLM \cite{bert} combined with their so called Translation Language Model (TLM), see Figure \ref{fig:mlm_tlm} for a comparison of the MLM and TLM (taken from Lample and Conneau).  

The TLM exploits a N-way parallel corpus of sentences to allow the model to explicitly use words from language A to predict the masked words in language B, hence encouraging the model to learn similar representations for semantically similar phrases across languages. The authors use a dataset accompanying the XNLI evaluation set which contains 10k parallel sentences in all 15 languages\footnote{The 15 languages in the corpus are English, French, German, Greek, Bulgarian, Russian, Turkish, Arabic, Vietnamese, Thai, Chinese, Hindi, Swahili and Urdu}. The TLM objective is altered with the MLM objective using Wikipedia dumps of each language. 

This approach differs from the one used by Artetxe and Schwenk, \shortcite{laser}: firstly, no encoder-decoder structure is used to explicitly align languages in a shared space. Instead, the model is only implicitly encouraged to align languages by allowing to share knowledge across the language boundary and solving the same task independently and simultaneously for both languages. Although the performance on the XNLI dataset improved using this method, there is an obvious drawback in terms of scaling to more languages due to the N-way parallel corpus requirement.


\subsection{Task-specific models}
All the above methods provide a means to extract word embeddings, which then serve as input to models for downstream tasks. 
We experiment with two downstream tasks: NER and POS tagging. These tasks are chosen in order to evaluate the performance on both semantic (NER) and syntactic level (POS tagging).

We use the encoder of the Transformer model as described in Vaswani et al. \shortcite{vaswani2017attention} as a sequence tagging model instead of a more commonly used RNN model in the hope to be able to transcend differences in sentence structure across languages. Specifically, we use a double layer Transformer with 2 attention-heads and 300 hidden dimensions for the query, key and value matrices as well as the feed forward network (FFN). The model is topped off by a Conditional Random Field (CRF) \cite{lafferty2001conditional}. For BERT and XLM, literature shows that adding a linear classification layer suffices for token-level classification tasks \cite{bert}.

\section{Experimental setup}
\subsection{Data and preprocessing}
\begin{table}[]
\resizebox{\linewidth}{!}{\begin{tabular}{l|llll}
                              &         & \textbf{LASER}             & \textbf{BERT}                 & \textbf{XLM}               \\
                              \hline
\multirow{2}{*}{\textbf{Supervised}}   & dataset & OPUS, 223M sentences & -                    & XNLI, 150k sents    \\
                              & task    & Translation          & -                    & TLM                 \\
\multirow{2}{*}{\textbf{Unsupervised}} & dataset & -                    & Wiki dump, 104 langs & Wiki dump, 15 langs \\
                              & task    & -                    & MLM + next sentence  & MLM                
\end{tabular}}
\caption{Pretraining data and tasks per architecture}
\label{table:pre_training_data}
\end{table}

We used the datasets from the CoNLL2002 \cite{conll2002} and CoNLL2003 \cite{conll2003} shared tasks, which provide data for the NER and POS in English, Spanish, Dutch and German. The data is gathered from local newspapers and is annotated with both named entities and POS tags. 
 All datasets are approximately the same size with $\pm15,000$ sentences to train on and $\pm3,500$ to test on. 

As the POS tags are given in language specific tags, we convert them to Universal POS tags \cite{petrov2011universal}, leaving us with 12 POS tags. In order to evaluate the ability of our methods to capture both semantic and syntactic information about the word no extra features are used to learn the model. 
All data is tokenized, and only punctuation normalization and lower casing has been applied in addition to that.
\subsection{Training}
\noindent \textbf{Baseline models} \qquad BPE BOW, BPE GRU and MUSE are relatively simple models and hence no intensive hyperparameter tuning is performed. Between the embedder and the sequence tagger a dropout of 0.25 is applied and within the sequence tagger a dropout of 0.15. We used Adam \cite{kingma2014adam} as an optimizer with the default learning rate of 0.001 and applied $L_2$ regularization with $\lambda = 0.001$. The models were trained for a total of 15 epochs while monitoring performance on a development set and applying early stopping \cite{EarlyStopping} after two rounds of consecutive decreased performance. \\

\noindent \textbf{LASER-based models} \qquad During preliminary experiments we found overfitting to be a major challenge for the LASER-based models and LSTM No Pretraining, hence more sophisticated techniques compared to the baselines have been applied for training. 

Firstly, instead of using Adam as optimizer, the \textit{1cycle LR}\cite{smith2018disciplined} policy is used. This policy uses the much simpler SGD optimizer with momentum and has been shown to improve generalization capabilities of neural networks while decreasing the number of epochs needed to train, a phenomenon the author calls "super convergence".

Finally, all remaining hyperparameters concerning regularization, such as dropout and $L_2$ regularization, are determined using Bayesian Optimization \cite{snoek2012practical}. \\

\noindent \textbf{Transformer-based models} \qquad As BERT and XLM are practically the same model except for their exact hidden size and their pretraining methods, they are trained using the same method. The original work \cite{bert} comes with a guide on how to finetune BERT for downstream NLP tasks based on a small grid search over values for the batch size, learning rate and number of epochs. Moreover, it contains optimal settings for the task of NER on the CoNLL2003 dataset, which is also part of our datasets. Apart from these specified optimal hyperparameter settings, the authors also note that BERT tends to be robust to the exact hyperparameter settings and hence we to use the specified hyperparameters for all experiments. This amounts to training for 4 epochs with a batch size of 16 using a learning rate of 5e-5.

\subsection{Experiments}
\noindent \textbf{Zero-shot transfer learning} \qquad The first set of experiments involve zero-shot transfer learning across languages. Each model is trained on the English dataset and consecutively evaluated on all other datasets, including two datasets in the low-resource languages --- Hungarian \cite{szarvas2006highly} and Basque \cite{alegria2004design} --- for NER. \\

\noindent \textbf{Joint training} \qquad In order to evaluate the benefit of joint training, we considered two scenarios. In the first scenario (\textbf{A}) a quarter of the training set of each language in the CoNLL2002 and CoNLL2003 shared tasks is taken and combined into a new training set of approximately the same size as the former ones. A validation set is created in a similar fashion and used for monitoring. Each model is trained on the multilingual dataset and then evaluated on the original test sets of each language. In the second scenario (\textbf{B}) one full training set (English) was complemented with a quarter of the training sets of the remaining languages. The difference in performance between scenarios A and B can be used as a way to quantify how well each model shares knowledge across languages.
\begin{table*}[!ht]
  
  \caption{F1-scores zero-shot transfer learning appended with monolingual scores if applicable}
  \label{tab:zero_transfer}
  \centering
  \resizebox{\textwidth}{!}{\begin{threeparttable}
    \begin{tabular}{l >{\raggedright}p{5em} l l l l l l l l}
      \toprule
      \tabhead{NER} &
      BPE BOW & BPE GRU &
      MUSE & LASER-top &
      LASER-elmo & LSTM No Pre & BERT
      & XLM\tnote{1}& Literature \\
      \midrule
      English & 0.509 & 0.742 & 0.758 & 0.803 & 0.786 & 0.752 & \textbf{0.914} & 0.892 & --/\underline{0.935}\\
      Dutch & 0.247\textit{/0.441} & 0.245\textit{/0.684} & 0.477\textit{/0.64} & 0.578\textit{/0.785} & 0.424\textit{/0.75} & 0.194\textit{/0.621} & \textbf{0.587\textit{/0.834}} & 0.212\textit{/0.792} & \underline{0.654}\\
      German & 0.187\textit{/0.55} & 0.242\textit{/0.57} & 0.481\textit{/0.603} & \underline{\textbf{0.617}}\textit{/0.679}  & 0.414\textit{/0.635} & 0.235\textit{/0.591}
          & 0.589\textit{/0.773} & 0.579\textit{/\textbf{0.813}} & 0.585 \\
      Spanish & 0.033\textit{/0.312} & 0.167\textit{/0.583} & 0.409\textit{/0.524} & 0.511\textit{/0.778} & 0.319\textit{/0.744} & 0.1567\textit{/0.635} &\textbf{0.618\textit{/0.819}} & 0.458\textit{/0.803} & \underline{0.735} \\
      Basque & 0.037 & 0.029 & ---\tnote{2} & 0.241  & 0.059 & 0.03 & \textbf{0.403} & 0.0811 &---\tnote{3}\\
    Hungarian & 0.023 & 0.043 & 0.387 & 0.414 & 0.208 & 0.073 &\textbf{0.455} & 0.13 &---\tnote{3}\\
      \toprule
      \tabhead{POS} \\
      \toprule
      English & 0.674 & 0.873 & 0.864 & 0.916 & 0.884 & 0.905 & \textbf{0.941} & 0.931 & ---\tnote{3} \\
      Dutch & 0.454\textit{/0.736}  & 0.430\textit{/0.931} & 0.563\textit{/0.923} & \textbf{0.726}\textit{/0.954} & 0.607\textit{/0.952} & 0.444\textit{/0.948} & 0.658\textit{/\textbf{0.961}} & 0.237\textit{/0.954} & ---\tnote{3}\\
      German & 0.456\textit{/0.79} & 0.466\textit{/0.912} & 0.623\textit{/0.895} & \textbf{0.781}\textit{/0.954} & 0.665\textit{/0.943} & 0.423/\textit{0.943} & 0.722\textit{/\textbf{0.958}} & 0.663\textit{/0.957} & ---\tnote{3}\\
      Spanish & 0.427\textit{/0.495} & 0.450\textit{/0.913} & 0.522\textit{/0.85} & 0.715\textit{/0.919} & 0.636\textit{/0.914} & 0.367\textit{/0.918} & \textbf{0.746\textit{/0.944}} & 0.601\textit{/0.941} & ---\tnote{3}\\
      \bottomrule
    \end{tabular}
    \begin{tablenotes}
      \item[1] XLM has been pretrained on the 15 XNLI languages: only English and German are amongst those languages.
      \item[2] No MUSE embeddings available for Basque.
      \item[3] Not available in literature.
    \end{tablenotes}
  \end{threeparttable}}
\end{table*}

\section{Results} \label{sec:results}

\subsection{Zero-shot transfer learning}
Table \ref{tab:zero_transfer} shows the F1-scores per model per language for the tasks of NER and POS tagging with the highest scores per language shown in bold and, if applicable, the state of the art underlined. Where possible, scores from monolingual training and evaluation are appended, separated by "/" for reference. All results have been tested against LASER-top for significance using the sign test with $\alpha =0.05$ and have been found to be significant.

 The highest performance scores were achieved by either BERT or LASER-top, where BERT performs the best on 6/9 tasks. Overall BERT appears to be a stronger model for learning the tasks at hand than the LSTM-based LASER-top model, as BERT achieves the highest scores in all monolingual settings except for German NER. LASER-top on the other hand is less capable of learning the task at hand in the source language, but the drop in performance when evaluating on other languages is smaller: it achieves the highest score on 2 out of 4 languages for POS tagging and advances the state of the art for German NER, indicating the added benefit of the LASER pretraining method for crosslingual knowledge sharing.

Surprisingly, XLM does not outperform BERT in any of the settings. It is worth noting that from the evaluated languages only English and German have been seen by XLM during pretraining, which explains the poor transfer learning capabilities to the remaining languages. Yet, XLM does not outperform BERT in the transfer from English to German, indicating no added benefit in the TLM method for zero-shot transfer learning across languages. 

For the low-resource languages the added benefit of contextualization is evident. As expected, performance on languages from more distant language-families is lower: all but one models score higher in Hungarian than in Basque. Furthermore, pretraining on a higher number of languages appears to positively influence performance on low-resource languages, as BERT outperforms XLM by a large margin and the same holds for LASER-top and LSTM No Pretraining.
 
Contrary to our expectations, LASER-top consistently outperforms LASER-elmo across all tasks. This is likely to be due to overfitting: LASER-elmo achieves higher scores on the training set than LASER-top in all experiments. Since the drop in performance across languages is far greater for all baseline models than for LASER-top and LASER-elmo, we attribute this improved performance to the multilingual pretraining. As the scores for LSTM No Pretraining are lower than LASER-top and LASER-elmo in the transferred languages this improved performance cannot be attributed to the added complexity from the extra layers.
\subsection{Joint training}
\begin{table*}[!h]
  \caption{F1-scores joint training}
  \label{tab:joint_training}
  \centering
  \resizebox{\textwidth}{!}{\begin{threeparttable}
    \resizebox{\textwidth}{!}{\begin{tabular}{l >{\raggedright}p{5em} l l l l l l}
      \toprule
      \tabhead{NER} &
      BPE BOW & BPE GRU &
      LASER-top & LASER-elmo & LSTM No Pre &
      BERT  & XLM\tnote{1}\\
      \midrule
      English & 0.330\textit{+0.096} & 0.632\textit{+0.017} & 0.723\textit{+0.012} & 0.678\textit{+0.011} & 0.617\textit{-0} & 0.826\textit{-0.038} & \textbf{0.854\textit{+0.001}} \\
      Dutch & 0.172\textit{+0.129}  & 0.527\textit{+0.071} & 0.709\textit{+0.03} & 0.618\textit{+0.045} & 0.601\textit{+0.012} & \textbf{0.804\textit{+0.017}} & 0.748\textit{+0.006}  \\
      German & 0.184\textit{+0.08} & 0.497\textit{+0.017} & 0.698\textit{+0.03} & 0.545\textit{+0.044} & 0.568\textit{+0.022} &
      0.799\textit{+0.012} & \textbf{0.809\textit{+0.022}} \\
      Spanish & 0.169\textit{+0.039} & 0.491\textit{+0.032} & 0.619\textit{+0.047} & 0.584\textit{+0.029} & 0.505\textit{+0.023} & \textbf{0.729}\textit{+0.005} & 0.724\textbf{\textit{+0.024}} \\
      Basque & 0.039\textit{+0.006} & 0.036\textit{-0.003} & 0.318\textit{-0.004}  & 0.131\textit{-0.047} & 0.04\textit{+0} & \textbf{0.502\textit{-0.}} & 0.144\textit{-0.013} \\
    Hungarian & 0.051\textit{+0.019} & 0.085\textit{+0.097} & 0.378\textit{+0.028} & 0.24\textit{+0.045} & 0.103\textit{-0} & \textbf{0.528\textit{+0.057}} & 0.254\textit{-0.024} \\
      \toprule 
      \tabhead{POS} \\
      \toprule
      English & 0.771\textit{+0.168} & 0.859\textit{-0.018} & 0.887\textit{+0.004} & 0.878\textit{-0.024} & 0.865\textit{-0.005} & 0.912\textit{+0.013}  & \textbf{0.916\textit{+0.002} } \\
      Dutch & 0.807\textit{+0.003}  & 0.91\textit{-0.006} & 0.927\textit{+0.006} & 0.919\textit{+0.007} & 0.912\textit{-0.005} & \textbf{0.949\textit{-0.}} & 0.928\textit{-0.003}  \\
      German & 0.884\textit{-.003} & 0.894\textit{-0.006} & 0.920\textit{+0.005} & 0.909\textit{+0.007} & 0.894\textit{-0.003} &
      0.937\textit{-0.002} & \textbf{0.94\textit{+0.005}} \\
      Spanish & 0.1\textit{-.035} & 0.876\textit{-0.006} & 0.9\textit{+0.004} & 0.886\textit{+0.006} & 0.879\textit{-0.003} & \textbf{0.915\textit{-0.004}} & 0.909\textit{+0.002} \\
      \bottomrule
    \end{tabular}}
    \begin{flushleft}
    \textbf{Note:} all results are depicted as base score followed by a deviation: the base scores are the F1 scores after training using a quarter of each dataset (scenario A) and the deviation is the change in scores after training with the full English train set and a quarter of the remaining languages (scenario B).
    \end{flushleft}
    \begin{tablenotes}
      \item[1] XLM has been pretrained on the 15 XNLI languages: only English and German are amongst those languages.
    \end{tablenotes}
  \end{threeparttable}}
\end{table*}

\begin{figure*}[!htbp]
\centering
\begin{minipage}{.5\linewidth}
  \centering
  \includegraphics[width=\linewidth, height=3cm]{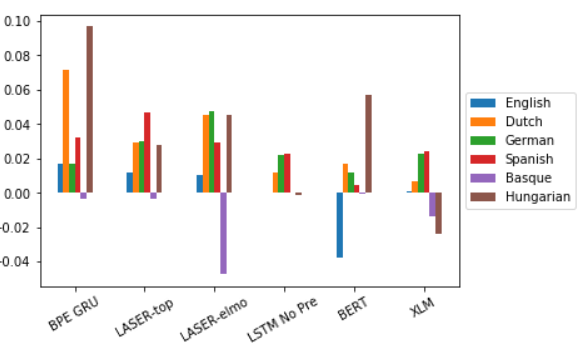}
  \captionof{figure}{Joint training NER}
  \label{fig:joint_training_ner}
\end{minipage}%
\begin{minipage}{0.5\linewidth}
  \centering
  \includegraphics[width=\linewidth, height=3cm]{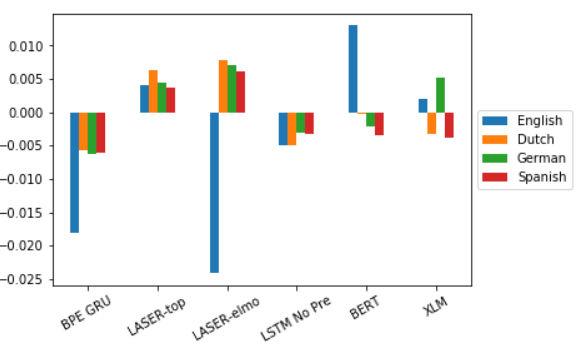}
  \captionof{figure}{Joint training POS}
  \label{fig:joint_training_pos}
\end{minipage}
\end{figure*}
Table \ref{tab:joint_training} shows the results of joint training of the four CoNLL2002 and CoNLL2003 languages. For both NER and POS tagging, BERT and XLM clearly outperform all other models, yet it is questionable whether this is the case because of the ability to share knowledge across languages or because Transformer-based models are better suited for the task. Figures \ref{fig:joint_training_ner} and \ref{fig:joint_training_pos} visualize the added benefit of joint training expressed as the difference in F1-scores compared to the baseline. For English this baseline is the monolingual baseline whereas for the other languages the baseline is the mixed setting with a quarter of each language (scenario A) compared to the full English dataset extended with a quarter of the remaining datasets (scenario B). BPE BOW has been omitted from the graphs as its values distort the graph and is of less importance than the remaining models. 

Whereas LASER-top benefits from joint training in all but one languages (Basque), this greatly differs for BERT and XLM, indicating that the pretraining method used for LASER might allow for better crosslingual knowledge sharing than the MLM and TLM methods used for BERT and XLM respectively. 

When comparing BERT and XLM, it appears that XLM shares knowledge better across languages it has been pretrained on while languages from a distant language family do not benefit at all from joint training. 

\subsection{Discussion and error analysis}
\begin{figure}[!ht]
\centering
\begin{subfigure}{0.5\linewidth}
  \centering
  \includegraphics[width=\linewidth]{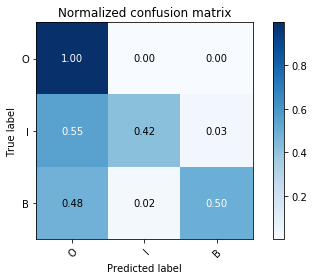}
\end{subfigure}%
\begin{subfigure}{0.5\linewidth}
  \centering
  \includegraphics[width=\linewidth]{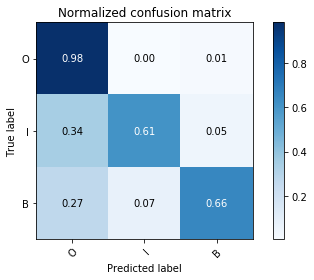}
\end{subfigure}
\caption{Dutch \textit{BIO} confusion for MUSE, LASER-top}
\label{fig:confusion_bio}
\end{figure}

In order to further analyze the added benefit of contextualized word embeddings versus static word embeddings in the zero-shot transfer setting, we compare MUSE with LASER-top in more detail. Firstly, we look at how good the models are at identifying whether an entity is present at all by trimming down the labels to \textit{B, I} and \textit{O} and creating a confusion matrix in Figure \ref{fig:confusion_bio}.
It can be clearly seen that LASER-top is better at detecting an entity than MUSE. 

Furthermore, we find evidence that the difference in performance is partially attributable to the fact that LASER-top has contextualized representations for words. Two examples are given where MUSE made an error that LASER-top did not by capitalizing the errors and appending the predicted entity type.
\begin{quote}
    \textit{despite winning the asian GAMES/O title two years ago , uzbekistan are in the finals as outsiders .} 
\end{quote}
\begin{quote}
    \textit{houston 1996-12-05 ohio state left tackle orlando PACE/O became the first repeat winner of the lombardi award thursday night when the ROTARY/O club of houston again honoured him as college football 's lineman of the year .}
\end{quote}
The incorrectly predicted words are all words that on their own would not be considered an entity of interest, but in this respective context they are part of a bigger entity span. These two sentences are examples of an often recurring pattern in the data in all evaluated languages.

\section{Conclusion}
In this work we have presented a comprehensive comparison of architectures and pretraining methods for contextualized multilingual word embeddings. We have also shown that it is possible to train a language model solely in an encoder-decoder style on the task of machine translation and consecutively use the encoder to create multilingual contextualized word embeddings. Moreover, we have shown that LASER-top outperforms (non-contextualized) baselines in multiple settings on multiple tasks and sometimes performs on par or better than BERT in the zero-shot transfer setting. 

Although our results indicate that our LSTM-based model is not as well suited for downstream NLP tasks as Transformer-based models, we have empirically shown our method to be superior at sharing knowledge across languages in a joint training setting and to perform at or above state of the art in zero-shot transfer setting.

As the results of our models are not yet on par with the current state of the art in monolingual settings, a logical next step is to investigate ways to combine pretraining methods, with the aim of learning higher quality monolingual word representations while encouraging knowledge sharing across languages. For instance a multi-task setup with BERT's MLM combined with the LASER pretraining method can be explored for this purpose. 

\section{ Acknowledgments}
This research was supported by Deloitte Risk Advisory B.V., NL. Special thanks to Willem Mobach, Tommie van der Bosch and Marc Verdonk for their involvement and support.

\bibliography{AAAI-HeijdenN.8664.bib}
\bibliographystyle{aaai}
\end{document}